\documentclass{article}

\PassOptionsToPackage{numbers, compress}{natbib}
\usepackage[dblblindworkshop,final]{neurips_2025}
\usepackage{amsmath}
\usepackage{graphicx}
\usepackage{subcaption}

\usepackage{listings}
\usepackage{xcolor}
\usepackage[T1]{fontenc}

\definecolor{codebg}{RGB}{249,249,240}
\definecolor{codekw}{RGB}{190,0,0}
\definecolor{codecm}{RGB}{0,128,0}
\definecolor{codeblt}{RGB}{0,102,204}
\definecolor{codety}{RGB}{163,21,21}

\lstdefinestyle{pyshade}{
  language=Python,
  backgroundcolor=\color{codebg},
  basicstyle=\ttfamily\small,
  keywordstyle=\color{codekw},
  commentstyle=\color{codecm},
  stringstyle=\color{codekw},
  numbers=none,
  frame=none,
  showstringspaces=false,
  keepspaces=true,
  columns=fullflexible,
  breaklines=true,
  aboveskip=6pt,
  belowskip=6pt,
  emph={len,range,zip,enumerate,map,filter,print,abs,min,max,sum},
  emphstyle=\color{codeblt},
  emph={[2]int,float,bool,bytes,tuple,list,dict,set},
  emphstyle={[2]\color{codety}},
}

\usepackage[utf8]{inputenc} 
\usepackage[T1]{fontenc}    
\usepackage{hyperref}       
\usepackage{url}            
\usepackage{booktabs}       
\usepackage{amsfonts}       
\usepackage{nicefrac}       
\usepackage{microtype}      
\usepackage{xcolor}         

\definecolor{diffplus}{RGB}{46,139,87}
\definecolor{diffminus}{RGB}{190,60,60}
\definecolor{diffmeta}{gray}{0.45}

\lstdefinelanguage{MyDiff}{
  morecomment=[f][\color{diffmeta}]{@@},
  morecomment=[f][\color{diffmeta}]{***},
  morecomment=[f][\color{diffmeta}]{---},
  morecomment=[f][\color{diffmeta}]{+++},
  morecomment=[f][\color{diffminus}]{-},
  morecomment=[f][\color{diffplus}]{+},
}

\lstdefinestyle{diffstyle}{
  style=pyshade,      
  language=MyDiff,
  numbers=none,
  showstringspaces=false,
  breaklines=true,
  frame=single
}

\title{Progress over Points: Reframing LM Benchmarks Around Scientific Objectives}
\workshoptitle{Evaluating the Evolving LLM Lifecycle: Benchmarks, Emergent Abilities, and Scaling}

\author{
  Alwin Jin\textsuperscript{1}\thanks{Work done while at Scale AI.} \quad
  Sean Hendryx\textsuperscript{2} \quad
  Vaskar Nath\textsuperscript{2} \\
  \textsuperscript{1}Georgia Institute of Technology \quad
  \textsuperscript{2}Scale AI \\
  \texttt{ajin37@gatech.edu}
}

\begin{document}

\maketitle

\begin{abstract}
Current benchmarks that test LLMs on static, already-solved problems (e.g., math word problems) effectively demonstrated basic capability acquisition. 
The natural progression has been toward larger, more comprehensive and challenging collections of static problems, an approach that inadvertently constrains the kinds of advances we can measure and incentivize.
To address this limitation, we argue for progress-oriented benchmarks, problem environments whose objectives are themselves the core targets of scientific progress, so that achieving state of the art on the benchmark \textit{advances the field}. As an introductory step, we instantiate an environment based on the NanoGPT speedrun. The environment standardizes a dataset slice, a reference model and training harness, and rich telemetry, with run-time verification and anti-gaming checks. Evaluation centers on the scientific delta achieved: best-attained loss and the efficiency frontier. Using this environment, we achieve a new state-of-the-art training time, improving upon the previous record by 3 seconds, and qualitatively observe the emergence of novel algorithmic ideas. Moreover, comparisons between models and agents remain possible, but they are a \textbf{means}, not the \textbf{end}; the benchmark’s purpose is to catalyze reusable improvements to the language modeling stack. With this release, the overarching goal is to seed a community shift from static problem leaderboards to test-time research on open-ended yet measurable scientific problems. In this new paradigm, progress on the benchmark is progress on the science, thus reframing "benchmarking" as a vehicle for scientific advancement. \footnote{Code available at \url{https://anonymous.4open.science/r/open-ended-benchmarks-private-DC26}}
\end{abstract}

\section{Introduction}

Progress in large language models (LLMs) is commonly tracked through static evaluations of already-solved problems: competition math and coding challenges, academic multiple-choice tests, and curated domain tasks such as reproducing known research results \cite{swe-bench, hle, paperbench, gsm8k}. While these benchmarks are important for evaluating whether models could master basic reasoning skills, they fail to evolve beyond measuring incremental improvements on closed-ended problems. This results in an approach that optimizes for narrow performance gains rather than meaningful scientific advancement.

As models and agentic systems approach capabilities for genuine scientific discovery, we argue for a fundamental shift toward \textbf{progress-oriented benchmarks}, where evaluation environments have objectives that are themselves targets of scientific progress. By adopting this new paradigm, performance gains become synonymous with advancing the field. Rather than awarding points for resolving pre-solved puzzles, these environments measure progress on important scientific objectives where ``doing well on the benchmark'' directly translates to scientific advancement.

We instantiate this paradigm through a standardized environment around the community NanoGPT speed-run \cite{modded_nanogpt_2024}, adapted for LLMs. The environment provides standardized data slices, reference models, training harnesses, and rich feedback metrics such as validation loss, profiling information, and training times, while also enforcing scientific integrity through anti-gaming protections. Evaluation then centers on the best-attained loss and movement of the efficiency frontier, shifting the focus to the \textbf{scientific delta achieved}. Although possible, model comparisons are used as diagnostic tools rather than ultimate objectives.


\paragraph{Why NanoGPT Speedrunning?} Language model pre-training represents a foundational area where algorithmic innovations routinely generalize beyond their initial tasks, from optimizer refinements to floating point precision changes and other architectural discoveries \cite{muon, lion, algorithmic-progress-lms}. The NanoGPT speedrun \cite{modded_nanogpt_2024} serves as a strong base for advancing pre-training techniques. This community-driven challenge searches for the fastest algorithm to train a language model to reach 3.28 cross-entropy validation loss on the FineWeb dataset, using a single 8xH100 GPU node. Since the first efforts in June 2024, the training time has steadily decreased from 45 minutes to just under 3 minutes. These advances were brought about by various algorithmic enhancements such as the Muon optimizer \cite{muon}. Discoveries improving NanoGPT training times often generalize to language modeling as a whole, making the speedrun an ideal environment for discovery.

\paragraph{Emphasizing Discovery while Allowing for Comparison.} Our environment allows for the plug-and-play of various systems, enabling different models or algorithms to be assessed on their ability to discover improvements. Surface-level comparisons can still be made by examining raw metric scores, but the environment also supports deeper, intermediary analysis. While model and agent comparisons remain valuable as diagnostic tools, lowering training times is still the final goal, thus framing benchmarking as a vehicle for scientific advancement instead of mere performance ranking.

The following summarizes the contributions of this work:
\begin{itemize}
    \item We reframe language model reasoning benchmarking from static problem scoring to measuring progress on scientific objectives, establishing principles for discovery-oriented evaluation environments.
    \item We present an open-ended evaluation environment based on the NanoGPT speedrun with standardized components and anti-gaming protections, preserving scientific integrity while promoting discovery.
    \item Through evolutionary test-time scaling, we demonstrate new state-of-the-art training times while providing diagnostics across various frontier models and agents, showing that the environment catalyzes reusable improvements beyond one-off leaderboard gains.
\end{itemize}

\section{Related Works}
\paragraph{Reasoning Benchmarking.}
Model and agent capabilities have increased rapidly in recent years, leading to the need to meaningfully evaluate their reasoning capabilities in fields such as math or science. Early work such as MMLU \cite{mmlu} evaluated broad academic and professional knowledge, but top models quickly reached saturation, prompting the development of harder, broader, and more reasoning-heavy benchmarks \cite{mmlu-pro, bbeh, agi-eval}. Domain-specific benchmarks with problem difficulties ranging from grade-school to graduate-level also emerged in math \cite{gsm8k, math-dataset, frontier-math}, science \cite{gpqa, olympiadbench}, and coding \cite{swe-bench, humaneval}. More recently, Humanity's Last Exam \cite{hle} assembled thousands of frontier problems across multiple subjects, positioning itself as the ``final'' academic benchmark for language models. However, even at this frontier of human knowledge, the paradigm of all aforementioned benchmarks remains static, where all problems have essentially been solved and solutions are closed-ended.

\paragraph{Open-Ended Benchmarking.} A growing line of work seeks to evaluate systems on more open-ended metrics. DSBench \cite{dsbench} and MLE-Bench \cite{mle-bench} assess agent capabilities on machine learning engineering tasks, such as hyperparameter tuning or code implementation, on constrained environments like Kaggle challenges. Although open-ended, scientific discovery is not emphasized as the evaluations have upper bounds on scores. Paperbench \cite{paperbench}, EXP-Bench \cite{exp-bench}, and the Automated LLM Speedrunning Benchmark \cite{automated-speedrunning} test the ability for systems to replicate prior scientific advancements. Given a research question and a brief description of methods, the system is evaluated on its ability to reproduce known results, with judging done using rubrics and LLM-as-judge. While science-centered, these benchmarks still fail to emphasize novel discovery. Other benchmarks such as LLM-SRBench \cite{llm-srbench} evaluate the equation discovery capabilities of models, requiring them to uncover symbolic relationships given a set of data. These relationships are typically constructed by rearranging existing equations and generating data under controlled settings, thus limiting the true scientific novelty of the evaluation. Nevertheless, open-ended benchmarks rarely evaluate on open-ended problems, especially in the reasoning realm.

\section{NanoGPT Evaluation Environment}
Motivated by open-ended, progress-oriented benchmarks, we introduce a rich evaluation environment centered around the NanoGPT speedrun \cite{modded_nanogpt_2024}.

\subsection{Evaluation Metrics}
Evaluation centers around two key metrics: cross-entropy validation loss and overall training time. Whereas the validation loss metrics primarily serves as a requirement check, achieving new bests on training time strongly implies the discovery of language modeling innovations.

The environment also captures incremental metrics. At designated training steps, we calculate the current step-averaged training time, the current number of iterations, as well as the current validation loss. This feedback can be used to inform of intermediary performance, or simply as additional calculated metrics after training is completed.

Although the aforementioned metrics enable more detailed performance tracking at an algorithmic level, they still lack fine-grained resource utilization feedback. To address this, we provide both profiling and hardware-level feedback in our NanoGPT environment. We profile key sections of the code---model forward pass, loss backward pass, optimizer steps, and data loading---tracking total run times, average run times, percentage of total time, and the number of calls for each one. At the hardware level, we capture the fifteen most time-consuming CUDA kernels and CPU operations along with their total call counts. Lastly, the environment provides overall training token throughput and peak memory usage amount. This suite of metrics provides a holistic picture of program performance, forming a rich reward surface for optimization.

\subsection{Core Algorithm}
Inspired by AlphEvolve-style evolution \cite{alphaevolve}, we fork and build on the open source OpenEvolve \cite{openevolve} implementation as the primary method to test and interact with our environment. The LLM system has four key components: a database $\mathbf{D}$, a program and prompt sampler $\mathbf{PS}$, an evaluator $\mathbf{E}$, and a language model $\mathbf{LM}$. For each evolutionary cycle $i$, a parent program $p_i$ is sampled from $\mathbf{D}$. The prompt sampler then creates a prompt involving $p_i$, which is used to prompt $\mathbf{LM}$. $\mathbf{LM}$ then suggests changes to $p_i$ to create child program $c_i$, which is evaluated by $\mathbf{E}$. Lastly, $c_i$ and its accompanying feedback metrics are stored in $\mathbf{D}$ and the next evolutionary cycle begins.

\textbf{Database}. The database maintains all programs and their associated metrics. In order to promote quality-diversity, we implement basic island evolution, where programs evolve within individual islands, and top perfomers periodically migrate between islands. The database serves to balance exploration and exploitation, and provides programs for inspiration sampling.

\textbf{Prompt Sampler}. The prompt sampler is responsible for both formatting prompts and sampling programs. As a baseline method, the sampler uses templates that are sampled from a pool of potential templates. Each prompt contains a top program set $T$ and a diverse program set $D$, where $|T|$ and $|D|$ are hyperparameters. Each top program $t_i \in T$ is sampled from the database. which maintains an archive of all elite programs across all islands. Each diverse program $d_i \in D$ is randomly sampled from the database from the set of non-elite programs $D \setminus T$. The prompt sampler compiles all of these programs with their associated metrics into a single prompt and asks for improvements. We then prompt for changes using unified search/replace blocks following aider.

\textbf{Meta-Prompting}. Although using static, templated prompts results in performance improvements, they struggle to elicit creative reasoning and idea generation. To address this, we implement a simple two-stage meta-prompting method. Given a prompt $P$ that already has program information formatted, we first ask the LM to provide a natural language solution $S$. The proposed solution $S$ is a high-level idea sketch emphasizing novelty and creativity. We then re-prompt the LM to implement $S$ using the same search/replace blocks as earlier. The key idea here is to decouple idea generation from code generation, thus reducing the cognitive load on the model and increasing creativity.

\textbf{Evaluator}. The evaluator executes child programs inside the environment, captures the rich feedback, and then updates the database accordingly. Our system also utilizes a fast error catcher where candidate programs are executed briefly to verify that they compile and run correctly. Erroneous programs are prompted for fixing by the LM before being fast-evaluated again. This cycle continues $N_\text{fast}$ times or until the program compiles, and is then sent to full evaluation. We also compute a singular score metric $s_c = t_\text{step} \cdot \ell_\text{val}$, where $t_\text{step}$ is the overall step average time and $\ell_\text{val}$ is the final validation loss. $s_c$ is used as the ultimate comparison metric within the database.

\begin{figure}
  \centering
  \includegraphics[width=\linewidth]{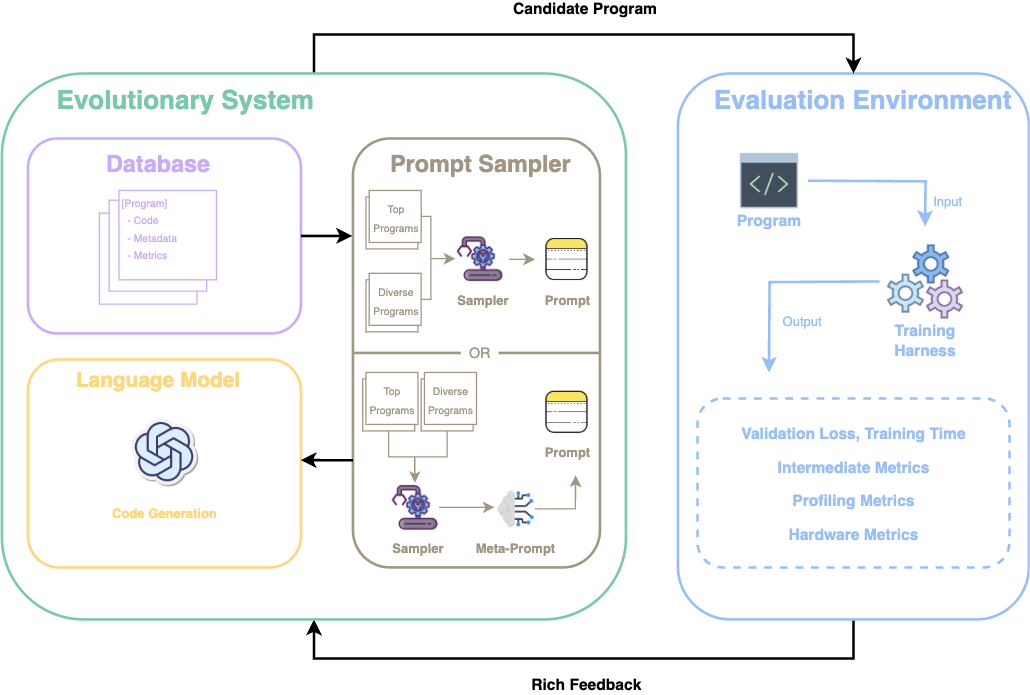}
  \caption{\textbf{Overview of our evaluation environment and the associated evolutionary system to interact with it}. The system stores prior programs in a database (purple). The prompt sampler (brown) then samples a child program and inspirations from the database, and uses various strategies to prompt the language model (yellow) for improvements. The resulting code is sent to the evaluation environment (blue), which executes the standardized training harness and returns rich feedback metrics that are stored in the database. }
\end{figure}

\textbf{Language Model}. The LM component of the system exposes a simple interface to plug-and-play various models.

\subsection{Anti-Gaming}
A significant challenge in providing a flexible optimization environment is its susceptibility to reward hacking, where trivial programs are discovered that maximize the reward signal but fail to achieve the intended goal. Our evaluation environment is explicitly designed to guard against these exploits.

\textbf{Runtime Injection of Evaluation Parameters}. Although the language model is free to modify any part of the code, we still guard against critical parameters. The environment injects key parameters during evaluation runtime to ensure integrity, overriding any earlier modifications. We enforce the exact training data slice, the validation slice, as well as the validation sequence length.

\textbf{Immutable Core Logic}. Another area where programs reward-hack is by changing core scientific logic. Our environment injects its own cross-entropy loss function, overriding the loss function implemented in the program. Although this restricts innovations involving loss calculation, the trade-off is necessary to prevent exploitation via trivial, non-equivalent loss functions. Future work could explore more robust guardrails such as verifying functional equivalence. The environment also prevents modification of the training document and causal masks. This prevents tokens from attending to prior tokens or to tokens from other documents.

\begin{figure*}[t]
  \centering

  \captionsetup[figure]{skip=12pt}
  \captionsetup[subfigure]{skip=10pt, belowskip=6pt}

  \begin{minipage}[c]{0.45\textwidth}
    \subcaptionbox{Comparison of top performers. Prolonged evolution within the evaluation environment yields human-level improvements. \label{fig:results_table}}{
      \resizebox{\linewidth}{!}{
        \begin{tabular}{|p{2.6cm}|c|c|}
          \hline
          \textbf{Program} & \textbf{Val. Loss} & \textbf{Time (s)} \\\hline
          \hline
          Previous SOTA & 3.280 & 176.7 \\
          Current SOTA  & \textbf{3.2797} & 175.2 \\
          Best Evolved  & 3.280 & \textbf{172.68} \\\hline
        \end{tabular}
      }
    }

    \vspace{0.5em}

    \subcaptionbox{Core optimization found in the best evolved program.\label{fig:code_snippet}}{
      \includegraphics[width=\linewidth]{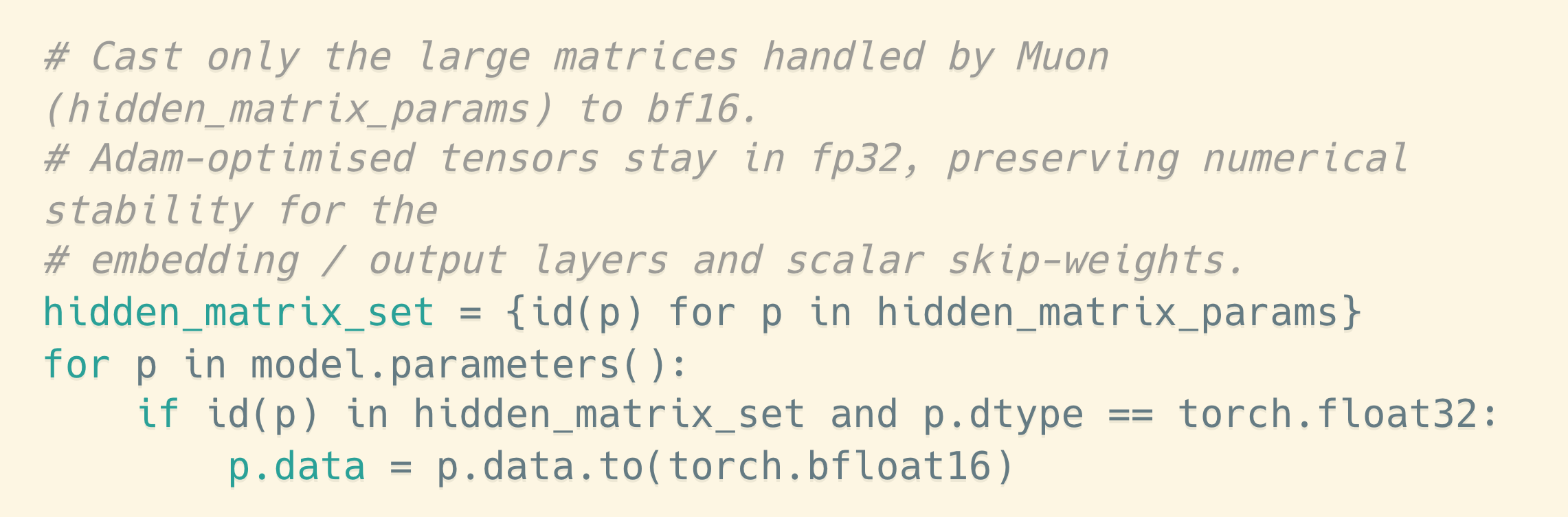}
    }
  \end{minipage}
  \hfill
  \begin{subfigure}[c]{0.50\textwidth}
    \centering
    \includegraphics[width=\linewidth]{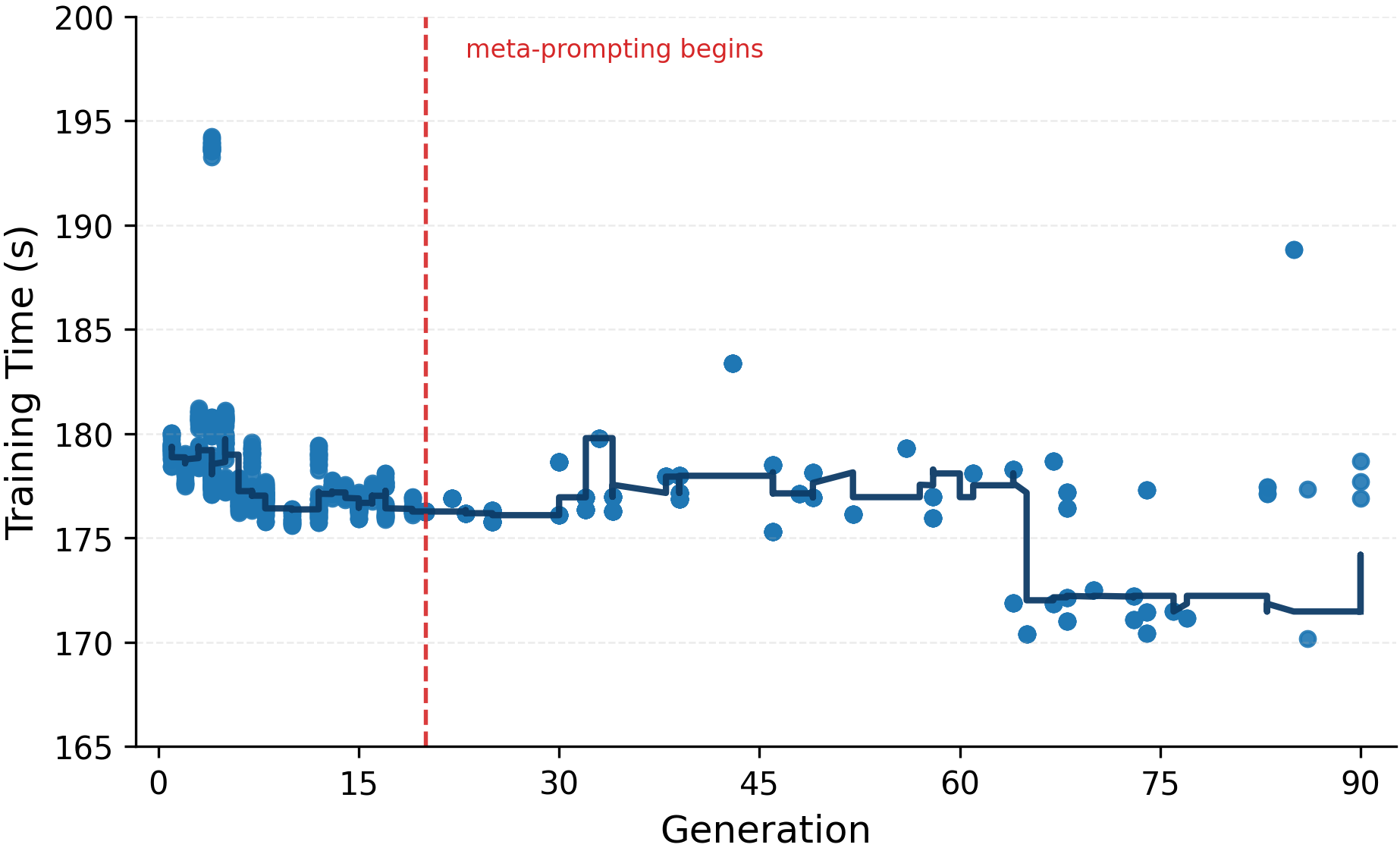}
    \caption{Performance of all programs that pass the validation loss threshold. We notice that meta-prompting enables state-of-art discovery but greatly reduces program success rate.}
    \label{fig:results-plot-large}
  \end{subfigure}
    \caption{Prolonged run of approximately 900 programs on the NanoGPT evaluation environment. The dotted red line in (c) marks the shift from stochastic templated prompting to meta-prompting.}
  \label{fig:combined_sota_fig}
\end{figure*}

\section{Experiments}
By framing the evaluation environment around scientific objectives, progress on the benchmark directly translates to progress in language modeling, whether at the algorithmic or optimization level. To this end, we investigate whether such evaluation environments can actually lead to discovery and new SOTA times by running experiments on various frontier models.

\subsection{Experimental Setup}
We experiment with our environment using various configurations of the modified OpenEvolve system. In order to scale up evolution, we generate and evaluate multiple children in parallel. The number of child programs created each iteration is termed the \textit{branching factor}. For all the following experiments, we use a branching factor of 10, a database elite archive size of 20, and $N_\text{fast} = 3$ fast retry attempts during evaluation. Because we notice meta-prompting often leads to more erroneous code, we disable meta-prompting for the first 20 iterations to build a strong parent pool of programs, relying solely on the stochastic template-based prompting.

\subsection{State-of-Art on NanoGPT}
We instantiate our system with the current (August 2025) leading NanoGPT code and perform a prolonged run using o3 \cite{o3} with 90 iterations. The results and details of the run are shown in Figure \ref{fig:combined_sota_fig}, and an intermediary analysis is given in section 4.3.

\textbf{Open-ended, environment-driven benchmarking shows potential for continuous discovery}. Table \ref{fig:results_table} highlights the benefits of shifting to this evaluation paradigm, as evolution with our NanoGPT environment yields a new SOTA time. The resulting core optimization is shown in Figure \ref{fig:code_snippet}, intelligently casting down the precision of optimizer operations on large hidden layer weight matrices. We see that the performance gain is non-trivial with the margin of improvement mirroring the previous human-set gains. Figure \ref{fig:results-plot-large} plots the training times of all programs that pass the validation loss threshold. We find that meta-prompting significantly degrades successful program rate, but greatly improves working program quality, and is thus important for state-of-art discovery. This suggests that our environment provides signals for discovery that stronger algorithms can meaningfully use, whether it be through stronger models, systems, or by scaling compute. Furthermore, the downwards trend line and low meta-prompt success rate emphasize the opportunity for improvements such as stronger models or deeper evolution runs, reinforcing the potential for continued discovery. Overall, shifting the benchmarking paradigm towards open-ended problems promotes progress towards advancing scientific fields.

\begin{figure*}[t]
  \centering

  \begin{subfigure}[b]{0.25\textwidth}
    \centering
    \resizebox{\linewidth}{!}{
      \begin{tabular}{|p{2.6cm}|c|}
        \hline
        \textbf{Model} & \textbf{Time (s)} \\\hline \hline
        GPT-5 Thinking  & 174.48 \\
        Gemini 2.5 Pro & 174.78 \\
        Claude 4 & \textbf{173.28} \\\hline
      \end{tabular}
    }
    \caption{Best found programs with acceptable loss. The reframed benchmarking paradigm still allows for comparisons.}
    \label{fig:comparison_a}
  \end{subfigure}
  \hfill
  \begin{subfigure}[b]{0.70\textwidth}
    \centering
    \includegraphics[width=\linewidth]{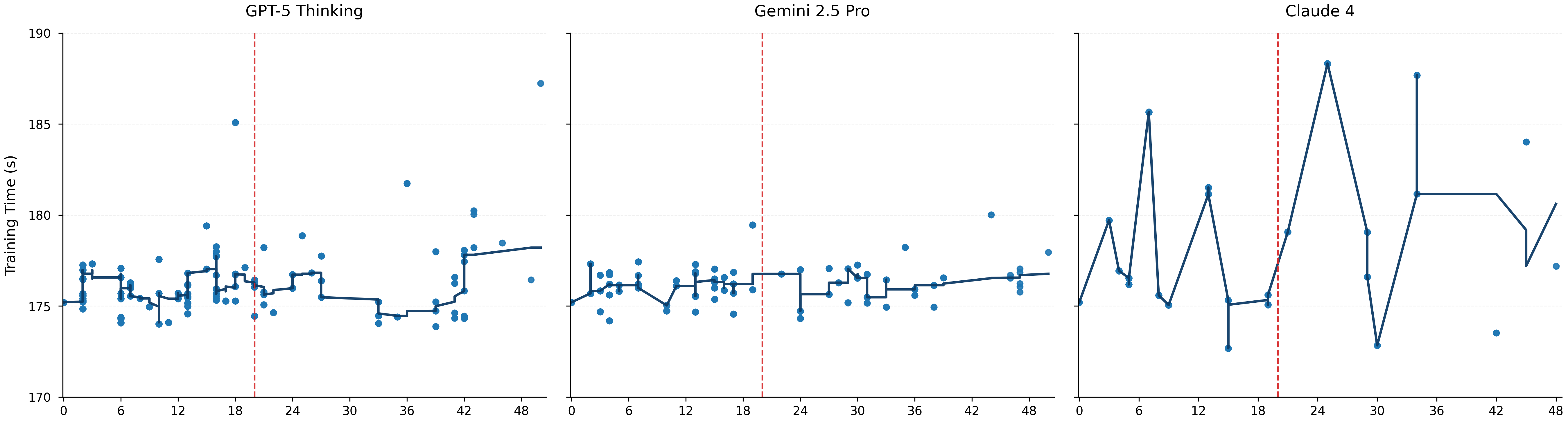}
    \caption{Performance of all acceptable programs on frontier models. The dotted red line represents when meta-prompting begins.}
    \label{fig:comparison_b}
  \end{subfigure}

  \par\medskip

  \begin{subfigure}[t]{\textwidth}
    \captionsetup{justification=raggedright,singlelinecheck=false}
    \centering
    \includegraphics[width=\linewidth]{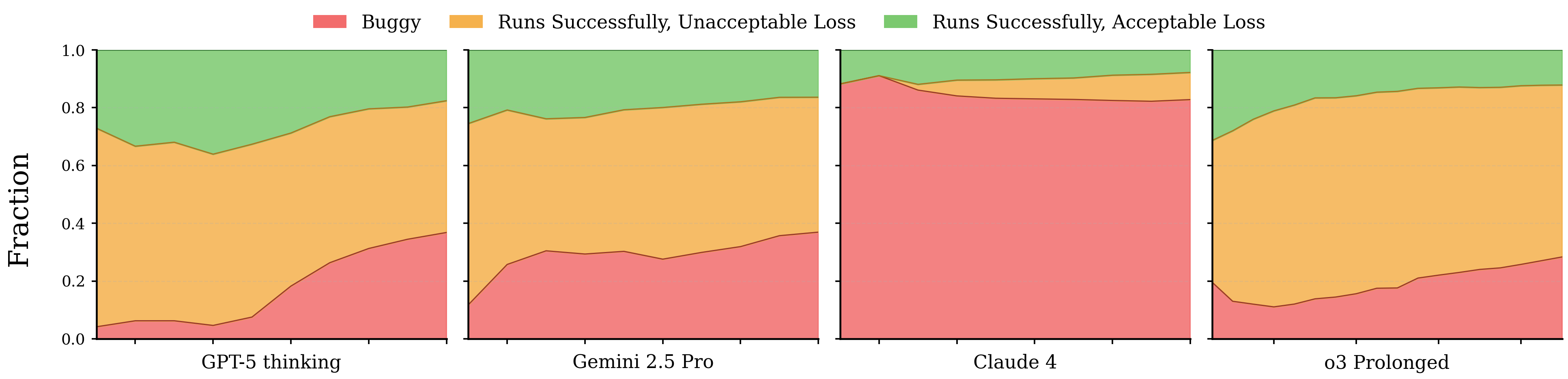}
    \caption{Breakdown of all programs found throughout evolution.}
    \label{fig:comparison_c}
  \end{subfigure}

  \caption{Our new benchmarking paradigm still allows for comparisons, but they are a means and not the end. The evaluation environment enables open-ended analysis of models and agents, offering insights beyond static numbers.}
  \label{fig:comparison_figure}
\end{figure*}

\subsection{Frontier Model Performance}
\label{sec:frontier-model-results}
Shifting to this new benchmarking paradigm still allows for the comparison of models and agents. Our environment and scaffolding support the plug-and-play of various models or even ensembles of models, enabling benchmarking comparisons in the traditional sense. However, these comparisons should serve as a means for introspection, with the true goal being faster training times. To that end, we run experiments on GPT-5 thinking \cite{gpt5}, Gemini 2.5 Pro \cite{gemini}, and Claude Sonnet 4 \cite{claude}. We initialize each model environment in the same manner as the o3 prolonged run, except only for 50 iterations. The results and some comparison analysis are provided in Figure \ref{fig:comparison_figure}

\textbf{Straightforward comparisons can be made using the primary optimization metric.} We can make quick comparisons across model, agent, or algorithm capabilities by examining the fastest final training times. From Table \ref{fig:comparison_a}, we find that Claude 4 yields the best time, and thus can be crowned the strongest in the traditional benchmarking sense. However, Figure \ref{fig:comparison_b} suggests that Claude struggles to consistently generate acceptable programs, and thus GPT-5 Thinking may perform better as we increase iterations due to its downwards trend line. Gemini still finds improvements from the current SOTA, but its trend line remains relatively flat throughout, suggesting a poorer ability to generate novel ideas or to integrate robust environmental feedback. These results highlight that comparisons serve as a strong means for analysis, but are not the end.

\begin{figure*}[!t]
  \centering

  \begin{subfigure}[t]{\textwidth}\vspace{0pt}
    \centering
    \includegraphics[height=0.42\textheight,keepaspectratio]{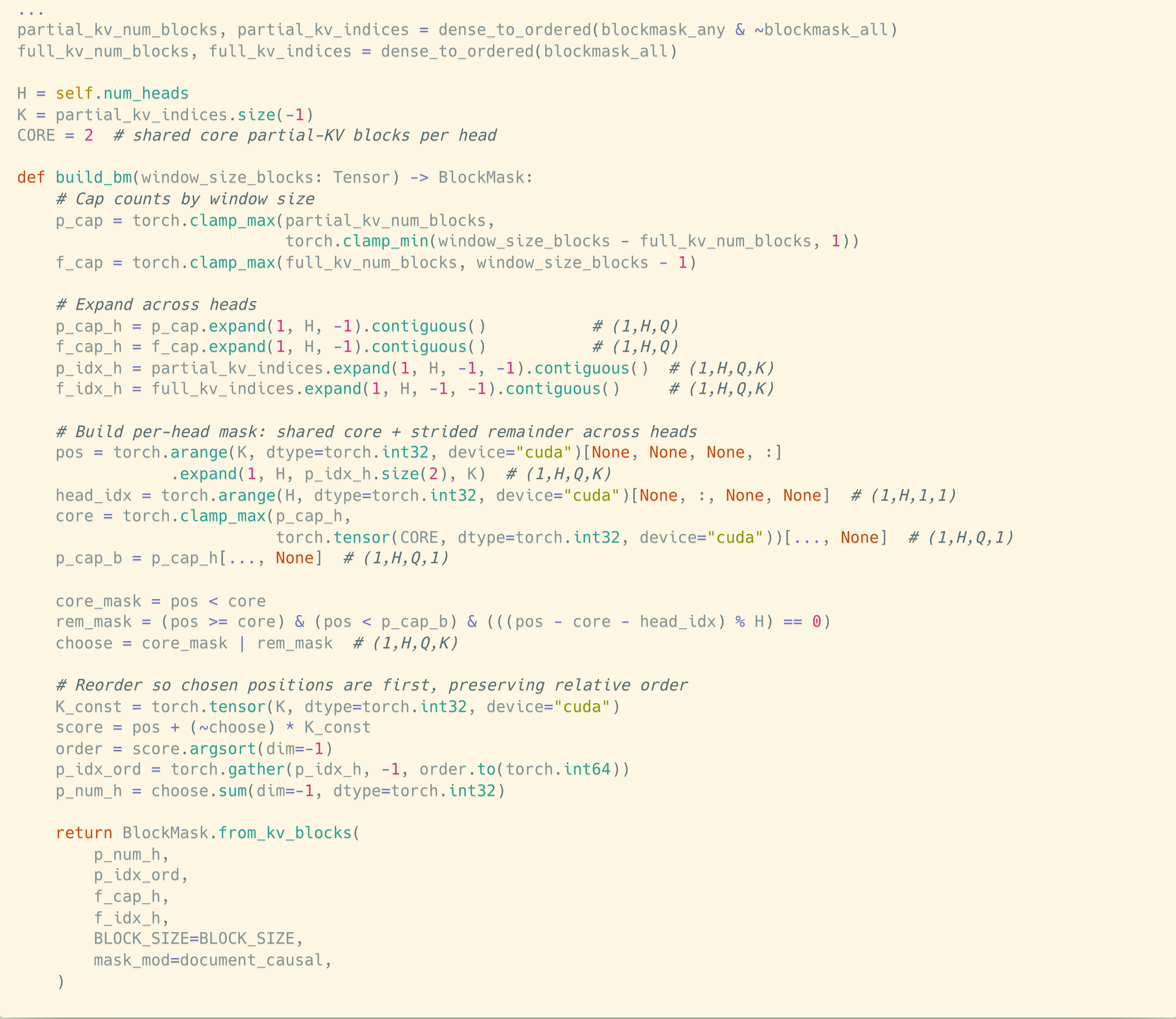}
    \caption{Sliding window attention optimization proposed by GPT-5.}
    \label{fig:interesting-programs-a}
  \end{subfigure}

  \par\smallskip

  \begin{subfigure}[t]{\textwidth}\vspace{0pt}
    \centering
    \includegraphics[height=0.42\textheight,keepaspectratio]{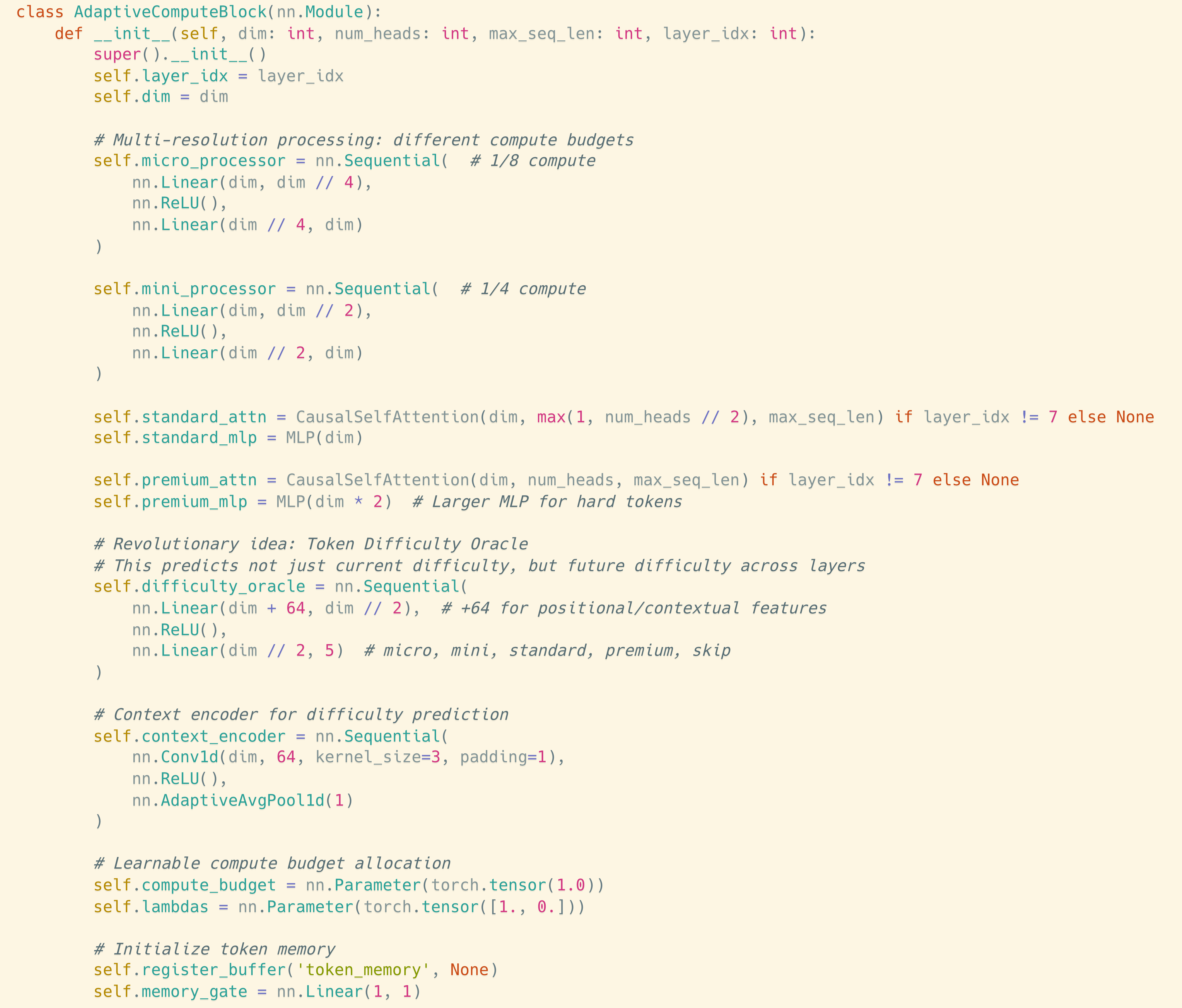}
    \caption{Token difficulty predictor and router proposed by Gemini.}
    \label{fig:interesting-programs-b}
  \end{subfigure}

  \caption{Interesting ideas emerge through evolution with open-ended evaluation environments.}
  \label{fig:interesting-programs}
\end{figure*}

\textbf{Evolution analysis.} Open-ended evaluation environments also allow for intermediary analysis of models and agents, which can be more valuable than static numbers. Figure \ref{fig:comparison_c} shows the fraction of programs in one of three categories at each evolution iteration. The red area denotes buggy programs, the orange represents non-buggy programs that have a validation loss too high for NanoGPT, and the green represents non-buggy, acceptable programs. Despite struggling greatly with producing working programs, Claude 4 still performs strongly, suggesting that improving its code generation capabilities could yield to strong discoveries. We also observe that after 20 iterations, all models see a spike in buggy program rate corresponding to the beginning of meta-prompting. Thus improving the meta-prompting and code generation capabilities of models is another promising avenue for improving performance on the benchmark.

\subsection{Emergence of Interesting Ideas}
A key phenomenon throughout evolution is the emergence of novel and interesting ideas. We present two of the most interesting concepts found by our system, both of which compile and execute successfully. Figure \ref{fig:interesting-programs-a} shows an innovative optimization to sliding window attention proposed by GPT-5. Instead of having each head repeat the same partial KV set, this new method keeps a small core of shared KV blocks every head attends to, and then distrubutes the remaining blocks in round-robin fashion amongst heads while maintaining relative order. This means that heads all attend to complementary ``stripes'' of KV blocks, resulting in the same coverage as earlier but with much fewer operations needed.

Figure \ref{fig:interesting-programs-b} was proposed by Gemini 2.5 and depicts an \texttt{AdaptiveComputeBlock}, a novel routing idea for transformers. This block contains a token difficulty predictor that use pooling to route tokens based on the current state and global context (the context being the surrounding tokens). These subpaths range from small MLPs to full attention layers. The \texttt{AdaptiveComputeBlock} also includes token memory and a learnable compute budget.

Although not groundbreaking by themselves, the emergence of these ideas throughout evolution highlights the potential of shifting to this open-ended evaluation paradigm.

\section{Future Work}
The results from our initial NanoGPT environment show the potential for open-ended evaluation environments to promote scientific discovery. The state-of-art speedrun result and emergence of interesting ideas during evolution help reinforce this idea. Key future directions include:
\begin{itemize}
    \item \textbf{Further scaling test-time compute.} Although our experiments used a non-trivial amount of test-time compute, scaling evolution to weeks or even months could lead to promising results. Future work could also explore additional frontier models (e.g. Grok or Claude Opus) as well as ensembling these models.
    \item \textbf{Creating more environments across reasoning domains.} The current NanoGPT environment primarily centers on pre-training advances. Similar environments extend naturally to post-training, with opportunities for discoveries in alignment and preference tuning. In math, environments would focus on unsolved problems, with progress to be measured through a combination of LLM-as-judge and formal theorem provers such as Lean.
\end{itemize}

\section{Conclusion}
In this work, we argue for a shift from static, puzzle-like benchmarking to progress-oriented, open-ended environments. These static evaluations have been invaluable for charting early progress in reasoning, but they increasingly fail to measure advances that materially improve the language–modeling stack. We demonstrate the potential of this new paradigm by instantiating a NanoGPT speedrun environment and run experiments using evolutionary test-time scaling methods on frontier models. These experiments produce a new state-of-the-art on the speedrun and qualitative analysis show the emergence of innovative algorithmic ideas, emphasizing the potential for continued scientific discovery.

\bibliographystyle{plainnat}
\bibliography{references}

\begin{thebibliography}{28}
\providecommand{\natexlab}[1]{#1}
\providecommand{\url}[1]{\texttt{#1}}
\expandafter\ifx\csname urlstyle\endcsname\relax
  \providecommand{\doi}[1]{doi: #1}\else
  \providecommand{\doi}{doi: \begingroup \urlstyle{rm}\Url}\fi

\bibitem[{Anthropic}(2025)]{claude}
{Anthropic}.
\newblock System card: Claude opus 4 \& claude sonnet 4, May 2025.
\newblock URL \url{https://www-cdn.anthropic.com/4263b940cabb546aa0e3283f35b686f4f3b2ff47.pdf}.
\newblock System card.

\bibitem[Chan et~al.(2024)Chan, Chowdhury, Jaffe, Aung, Sherburn, Mays, Starace, Liu, Maksin, Patwardhan, Weng, and Mądry]{mle-bench}
Jun~Shern Chan, Neil Chowdhury, Oliver Jaffe, James Aung, Dane Sherburn, Evan Mays, Giulio Starace, Kevin Liu, Leon Maksin, Tejal Patwardhan, Lilian Weng, and Aleksander Mądry.
\newblock Mle-bench: Evaluating machine learning agents on machine learning engineering.
\newblock 2024.
\newblock URL \url{https://arxiv.org/abs/2410.07095}.

\bibitem[Chen et~al.(2021)Chen, Tworek, Jun, Yuan, de~Oliveira~Pinto, Kaplan, Edwards, Burda, Joseph, Brockman, Ray, Puri, Krueger, Petrov, Khlaaf, Sastry, Mishkin, Chan, Gray, Ryder, Pavlov, Power, Kaiser, Bavarian, Winter, Tillet, Such, Cummings, Plappert, Chantzis, Barnes, Herbert-Voss, Guss, Nichol, Paino, Tezak, Tang, Babuschkin, Balaji, Jain, Saunders, Hesse, Carr, Leike, Achiam, Misra, Morikawa, Radford, Knight, Brundage, Murati, Mayer, Welinder, McGrew, Amodei, McCandlish, Sutskever, and Zaremba]{humaneval}
Mark Chen, Jerry Tworek, Heewoo Jun, Qiming Yuan, Henrique~Ponde de~Oliveira~Pinto, Jared Kaplan, Harri Edwards, Yuri Burda, Nicholas Joseph, Greg Brockman, Alex Ray, Raul Puri, Gretchen Krueger, Michael Petrov, Heidy Khlaaf, Girish Sastry, Pamela Mishkin, Brooke Chan, Scott Gray, Nick Ryder, Mikhail Pavlov, Alethea Power, Lukasz Kaiser, Mohammad Bavarian, Clemens Winter, Philippe Tillet, Felipe~Petroski Such, Dave Cummings, Matthias Plappert, Fotios Chantzis, Elizabeth Barnes, Ariel Herbert-Voss, William~Hebgen Guss, Alex Nichol, Alex Paino, Nikolas Tezak, Jie Tang, Igor Babuschkin, Suchir Balaji, Shantanu Jain, William Saunders, Christopher Hesse, Andrew~N. Carr, Jan Leike, Josh Achiam, Vedant Misra, Evan Morikawa, Alec Radford, Matthew Knight, Miles Brundage, Mira Murati, Katie Mayer, Peter Welinder, Bob McGrew, Dario Amodei, Sam McCandlish, Ilya Sutskever, and Wojciech Zaremba.
\newblock Evaluating large language models trained on code.
\newblock 2021.

\bibitem[Chen et~al.(2023)Chen, Liang, Huang, Real, Wang, Liu, Pham, Dong, Luong, Hsieh, Lu, and Le]{lion}
Xiangning Chen, Chen Liang, Da~Huang, Esteban Real, Kaiyuan Wang, Yao Liu, Hieu Pham, Xuanyi Dong, Thang Luong, Cho-Jui Hsieh, Yifeng Lu, and Quoc~V. Le.
\newblock Symbolic discovery of optimization algorithms, 2023.
\newblock URL \url{https://arxiv.org/abs/2302.06675}.

\bibitem[Cobbe et~al.(2021)Cobbe, Kosaraju, Bavarian, Chen, Jun, Kaiser, Plappert, Tworek, Hilton, Nakano, Hesse, and Schulman]{gsm8k}
Karl Cobbe, Vineet Kosaraju, Mohammad Bavarian, Mark Chen, Heewoo Jun, Lukasz Kaiser, Matthias Plappert, Jerry Tworek, Jacob Hilton, Reiichiro Nakano, Christopher Hesse, and John Schulman.
\newblock Training verifiers to solve math word problems, 2021.
\newblock URL \url{https://arxiv.org/abs/2110.14168}.

\bibitem[Glazer et~al.(2025)Glazer, Erdil, Besiroglu, Chicharro, Chen, Gunning, Olsson, Denain, Ho, de~Oliveira~Santos, Järviniemi, Barnett, Sandler, Vrzala, Sevilla, Ren, Pratt, Levine, Barkley, Stewart, Grechuk, Grechuk, Enugandla, and Wildon]{frontier-math}
Elliot Glazer, Ege Erdil, Tamay Besiroglu, Diego Chicharro, Evan Chen, Alex Gunning, Caroline~Falkman Olsson, Jean-Stanislas Denain, Anson Ho, Emily de~Oliveira~Santos, Olli Järviniemi, Matthew Barnett, Robert Sandler, Matej Vrzala, Jaime Sevilla, Qiuyu Ren, Elizabeth Pratt, Lionel Levine, Grant Barkley, Natalie Stewart, Bogdan Grechuk, Tetiana Grechuk, Shreepranav~Varma Enugandla, and Mark Wildon.
\newblock Frontiermath: A benchmark for evaluating advanced mathematical reasoning in ai, 2025.
\newblock URL \url{https://arxiv.org/abs/2411.04872}.

\bibitem[Google(2025)]{gemini}
Google.
\newblock Gemini 2.5: Pushing the frontier with advanced reasoning, multimodality, long context, and next generation agentic capabilities, 2025.
\newblock URL \url{https://arxiv.org/abs/2507.06261}.

\bibitem[He et~al.(2024)He, Luo, Bai, Hu, Thai, Shen, Hu, Han, Huang, Zhang, Liu, Qi, Liu, and Sun]{olympiadbench}
Chaoqun He, Renjie Luo, Yuzhuo Bai, Shengding Hu, Zhen~Leng Thai, Junhao Shen, Jinyi Hu, Xu~Han, Yujie Huang, Yuxiang Zhang, Jie Liu, Lei Qi, Zhiyuan Liu, and Maosong Sun.
\newblock Olympiadbench: A challenging benchmark for promoting agi with olympiad-level bilingual multimodal scientific problems, 2024.

\bibitem[Hendrycks et~al.(2021{\natexlab{a}})Hendrycks, Burns, Basart, Zou, Mazeika, Song, and Steinhardt]{mmlu}
Dan Hendrycks, Collin Burns, Steven Basart, Andy Zou, Mantas Mazeika, Dawn Song, and Jacob Steinhardt.
\newblock Measuring massive multitask language understanding, 2021{\natexlab{a}}.
\newblock URL \url{https://arxiv.org/abs/2009.03300}.

\bibitem[Hendrycks et~al.(2021{\natexlab{b}})Hendrycks, Burns, Kadavath, Arora, Basart, Tang, Song, and Steinhardt]{math-dataset}
Dan Hendrycks, Collin Burns, Saurav Kadavath, Akul Arora, Steven Basart, Eric Tang, Dawn Song, and Jacob Steinhardt.
\newblock Measuring mathematical problem solving with the math dataset.
\newblock \emph{NeurIPS}, 2021{\natexlab{b}}.

\bibitem[Ho et~al.(2024)Ho, Besiroglu, Erdil, Owen, Rahman, Guo, Atkinson, Thompson, and Sevilla]{algorithmic-progress-lms}
Anson Ho, Tamay Besiroglu, Ege Erdil, David Owen, Robi Rahman, Zifan~Carl Guo, David Atkinson, Neil Thompson, and Jaime Sevilla.
\newblock Algorithmic progress in language models, 2024.
\newblock URL \url{https://arxiv.org/abs/2403.05812}.

\bibitem[Jimenez et~al.(2024)Jimenez, Yang, Wettig, Yao, Pei, Press, and Narasimhan]{swe-bench}
Carlos~E Jimenez, John Yang, Alexander Wettig, Shunyu Yao, Kexin Pei, Ofir Press, and Karthik~R Narasimhan.
\newblock {SWE}-bench: Can language models resolve real-world github issues?
\newblock In \emph{The Twelfth International Conference on Learning Representations}, 2024.
\newblock URL \url{https://openreview.net/forum?id=VTF8yNQM66}.

\bibitem[Jing et~al.(2024)Jing, Huang, Wang, Yao, Yu, Ma, Zhang, Du, and Yu]{dsbench}
Liqiang Jing, Zhehui Huang, Xiaoyang Wang, Wenlin Yao, Wenhao Yu, Kaixin Ma, Hongming Zhang, Xinya Du, and Dong Yu.
\newblock Dsbench: How far are data science agents to becoming data science experts?, 2024.
\newblock URL \url{https://arxiv.org/abs/2409.07703}.

\bibitem[Jordan et~al.(2024{\natexlab{a}})Jordan, Bernstein, Rappazzo, @fernbear.bsky.social, Vlado, Jiacheng, Cesista, Koszarsky, and @Grad62304977]{modded_nanogpt_2024}
Keller Jordan, Jeremy Bernstein, Brendan Rappazzo, @fernbear.bsky.social, Boza Vlado, You Jiacheng, Franz Cesista, Braden Koszarsky, and @Grad62304977.
\newblock modded-nanogpt: Speedrunning the nanogpt baseline, 2024{\natexlab{a}}.
\newblock URL \url{https://github.com/KellerJordan/modded-nanogpt}.

\bibitem[Jordan et~al.(2024{\natexlab{b}})Jordan, Jin, Boza, Jiacheng, Cesista, Newhouse, and Bernstein]{muon}
Keller Jordan, Yuchen Jin, Vlado Boza, You Jiacheng, Franz Cesista, Laker Newhouse, and Jeremy Bernstein.
\newblock Muon: An optimizer for hidden layers in neural networks, 2024{\natexlab{b}}.
\newblock URL \url{https://kellerjordan.github.io/posts/muon/}.

\bibitem[Kazemi et~al.(2025)Kazemi, Fatemi, Bansal, Palowitch, Anastasiou, Mehta, Jain, Aglietti, Jindal, Chen, Dikkala, Tyen, Liu, Shalit, Chiappa, Olszewska, Tay, Tran, Le, and Firat]{bbeh}
Mehran Kazemi, Bahare Fatemi, Hritik Bansal, John Palowitch, Chrysovalantis Anastasiou, Sanket~Vaibhav Mehta, Lalit~K. Jain, Virginia Aglietti, Disha Jindal, Peter Chen, Nishanth Dikkala, Gladys Tyen, Xin Liu, Uri Shalit, Silvia Chiappa, Kate Olszewska, Yi~Tay, Vinh~Q. Tran, Quoc~V. Le, and Orhan Firat.
\newblock Big-bench extra hard, 2025.
\newblock URL \url{https://arxiv.org/abs/2502.19187}.

\bibitem[Kon et~al.(2024)Kon, Liu, Zhu, Ding, Peng, Xing, Huang, Qiu, Srinivasa, Lee, Chowdhury, Zaharia, and Chen]{exp-bench}
Patrick Tser~Jern Kon, Jiachen Liu, Xinyi Zhu, Qiuyi Ding, Jingjia Peng, Jiarong Xing, Yibo Huang, Yiming Qiu, Jayanth Srinivasa, Myungjin Lee, Mosharaf Chowdhury, Matei Zaharia, and Ang Chen.
\newblock Exp-bench: Can ai conduct ai research experiments?
\newblock 2024.
\newblock URL \url{https://arxiv.org/abs/2410.07095}.

\bibitem[Novikov et~al.(2025)Novikov, Vũ, Eisenberger, Dupont, Huang, Wagner, Shirobokov, Kozlovskii, Ruiz, Mehrabian, Kumar, See, Chaudhuri, Holland, Davies, Nowozin, Kohli, and Balog]{alphaevolve}
Alexander Novikov, Ngân Vũ, Marvin Eisenberger, Emilien Dupont, Po-Sen Huang, Adam~Zsolt Wagner, Sergey Shirobokov, Borislav Kozlovskii, Francisco J.~R. Ruiz, Abbas Mehrabian, M.~Pawan Kumar, Abigail See, Swarat Chaudhuri, George Holland, Alex Davies, Sebastian Nowozin, Pushmeet Kohli, and Matej Balog.
\newblock Alphaevolve: A coding agent for scientific and algorithmic discovery, 2025.
\newblock URL \url{https://arxiv.org/abs/2506.13131}.

\bibitem[{OpenAI}(2025{\natexlab{a}})]{gpt5}
{OpenAI}.
\newblock Gpt-5 system card.
\newblock System card, OpenAI, August 2025{\natexlab{a}}.
\newblock URL \url{https://cdn.openai.com/gpt-5-system-card.pdf}.

\bibitem[{OpenAI}(2025{\natexlab{b}})]{o3}
{OpenAI}.
\newblock Openai o3 and o4-mini system card.
\newblock Technical report, OpenAI, 2025{\natexlab{b}}.
\newblock URL \url{https://openai.com/index/o3-o4-mini-system-card}.

\bibitem[Phan et~al.(2025)Phan, Gatti, Han, Li, Hu, Zhang, Zhang, Shaaban, Ling, Shi, Choi, Agrawal, Chopra, Khoja, Kim, Ren, Hausenloy, Zhang, Mazeika, Dodonov, Nguyen, Lee, Anderson, Doroshenko, Stokes, Mahmood, Pokutnyi, Iskra, Wang, Levin, Kazakov, Feng, Feng, Zhao, Yu, Gangal, Zou, Wang, Popov, Gerbicz, Galgon, Schmitt, Yeadon, Lee, Sauers, Sanchez, Giska, Roth, Riis, Utpala, Burns, Goshu, Naiya, Agu, Giboney, Cheatom, Fournier-Facio, Crowson, Finke, Cheng, Zampese, Hoerr, Nandor, Park, Gehrunger, Cai, McCarty, Garretson, Taylor, Sileo, Ren, Qazi, Li, Nam, Wydallis, Arkhipov, Shi, Bacho, Willcocks, Cao, Motwani, de~Oliveira~Santos, Veith, Vendrow, Cojoc, Zenitani, Robinson, Tang, Li, Vendrow, Fraga, Kuchkin, Maksimov, Marion, Efremov, Lynch, Liang, Mikov, Gritsevskiy, Guillod, Demir, Martinez, Pageler, Zhou, Soori, Press, Tang, Rissone, Green, Brüssel, Twayana, Dieuleveut, Imperial, Prabhu, Yang, Crispino, Rao, Zvonkine, Loiseau, Kalinin, Lukas, Manolescu, Stambaugh, Mishra, Hogg, Bosio, Coppola,
  Salazar, Jin, Sayous, Ivanov, Schwaller, Senthilkuma, Bran, Algaba, den Houte, Sypt, Verbeken, Noever, Kopylov, Myklebust, Li, Schut, Zheltonozhskii, Yuan, Lim, Stanley, Yang, Maar, Wykowski, Oller, Sahu, Ardito, Hu, Kamdoum, Jin, Vilchis, Zu, Lackner, Koppel, Sun, Antonenko, Chern, Zhao, Arsene, Cavanagh, Li, Shen, Crisostomi, Zhang, Dehghan, Ivanov, Perrella, Kaparov, Zang, Sucholutsky, Kharlamova, Orel, Poritski, Ben-David, Berger, Whitfill, Foster, Munro, Ho, Sivarajan, Hava, Kuchkin, Holmes, Rodriguez-Romero, Sommerhage, Zhang, Moat, Schneider, Kazibwe, Clarke, Kim, Dias, Fish, Elser, Kreiman, Vilchis, Klose, Anantheswaran, Zweiger, Rawal, Li, Nguyen, Daans, Heidinger, Radionov, Rozhoň, Ginis, Stump, Cohen, Poświata, Tkadlec, Goldfarb, Wang, Padlewski, Barzowski, Montgomery, Stendall, Tucker-Foltz, Stade, Rogers, Goertzen, Grabb, Shukla, Givré, Ambay, Sen, Aziz, Inlow, He, Zhang, Kaddar, Ängquist, Chen, Wang, Ramakrishnan, Thornley, Terpin, Schoelkopf, Zheng, Carmi, Brown, Zhu, Bartolo, Wheeler,
  Stehberger, Bradshaw, Heimonen, Sridhar, Akov, Sandlin, Makarychev, Tam, Hoang, Cunningham, Goryachev, Patramanis, Krause, Redenti, Aldous, Lai, Coleman, Xu, Lee, Magoulas, Zhao, Tang, Cohen, Paradise, Kirchner, Ovchynnikov, Matos, Shenoy, Wang, Nie, Sztyber-Betley, Faraboschi, Riblet, Crozier, Halasyamani, Verma, Joshi, Meril, Ma, Andréoletti, Singhal, Platnick, Nevirkovets, Basler, Ivanov, Khoury, Gustafsson, Piccardo, Mostaghimi, Chen, Singh, Khánh, Rosu, Szlyk, Brown, Narayan, Menezes, Roberts, Alley, Sun, Patel, Lamparth, Reuel, Xin, Xu, Loader, Martin, Wang, Achilleos, Preu, Korbak, Bosio, Kazemi, Chen, Bálint, Lo, Wang, Nunes, Milbauer, Bari, Wang, Ansarinejad, Sun, Durand, Elgnainy, Douville, Tordera, Balabanian, Wolff, Kvistad, Milliron, Sakor, Eron, O., Shah, Zhou, Kamalov, Abdoli, Santens, Barkan, Tee, Zhang, Tomasiello, Luca, Looi, Le, Kolt, Pan, Rodman, Drori, Fossum, Muennighoff, Jagota, Pradeep, Fan, Eicher, Chen, Thaman, Merrill, Firsching, Harris, Ciobâcă, Gross, Pandey, Gusev, Jones,
  Agnihotri, Zhelnov, Mofayezi, Piperski, Zhang, Dobarskyi, Leventov, Soroko, Duersch, Taamazyan, Ho, Ma, Held, Xian, Zebaze, Mohamed, Leser, Yuan, Yacar, Lengler, Olszewska, Fratta, Oliveira, Jackson, Zou, Chidambaram, Manik, Haffenden, Stander, Dasouqi, Shen, Golshani, Stap, Kretov, Uzhou, Zhidkovskaya, Winter, Rodriguez, Lauff, Wehr, Tang, Hossain, Phillips, Samuele, Ekström, Hammon, Patel, Farhidi, Medley, Mohammadzadeh, Peñaflor, Kassahun, Friedrich, Perez, Pyda, Sakal, Dhamane, Mirabadi, Hallman, Okutsu, Battaglia, Maghsoudimehrabani, Amit, Hulbert, Pereira, Weber, Handoko, Peristyy, Malina, Mehkary, Aly, Reidegeld, Dick, Friday, Singh, Shapourian, Kim, Costa, Gurdogan, Kumar, Ceconello, Zhuang, Park, Carroll, Tawfeek, Steinerberger, Aggarwal, Kirchhof, Dai, Kim, Ferret, Shah, Wang, Yan, Burdzy, Zhang, Franca, Pham, Loh, Robinson, Jackson, Giordano, Petersen, Cosma, Colino, White, Votava, Vinnikov, Delaney, Spelda, Stritecky, Shahid, Mourrat, Vetoshkin, Sponselee, Bacho, Yong, de~la Rosa, Cho, Li,
  Malod, Weller, Albani, Lang, Laurendeau, Kazakov, Adesanya, Portier, Hollom, Souza, Zhou, Degorre, Yalın, Obikoya, Rai, Bigi, Boscá, Shumar, Bacho, Recchia, Popescu, Shulga, Tanwie, Lux, Rank, Ni, Brooks, Yakimchyk, Huanxu, Liu, Cavalleri, Häggström, Verkama, Newbould, Gundlach, Brito-Santana, Amaro, Vajipey, Grover, Wang, Kratish, Li, Gopi, Caciolai, de~Witt, Hernández-Cámara, Rodolà, Robins, Williamson, Cheng, Raynor, Qi, Segev, Fan, Martinson, Wang, Hausknecht, Brenner, Mao, Demian, Kassani, Zhang, Avagian, Scipio, Ragoler, Tan, Sims, Plecnik, Kirtland, Bodur, Shinde, Labrador, Adoul, Zekry, Karakoc, Santos, Shamseldeen, Karim, Liakhovitskaia, Resman, Farina, Gonzalez, Maayan, Anderson, Pena, Kelley, Mariji, Pouriamanesh, Wu, Finocchio, Alarab, Cole, Ferreira, Johnson, Safdari, Dai, Arthornthurasuk, McAlister, Moyano, Pronin, Fan, Ramirez-Trinidad, Malysheva, Pottmaier, Taheri, Stepanic, Perry, Askew, Rodríguez, Minissi, Lorena, Iyer, Fasiludeen, Clark, Ducey, Piza, Somrak, Vergo, Qin, Borbás,
  Chu, Lindsey, Jallon, McInnis, Chen, Semler, Gloor, Shah, Carauleanu, Lauer, Đuc Huy, Shahrtash, Duc, Lewark, Brown, Albanie, Weber, Vaz, Clavier, Fan, e~Silva, Long, Lian, Abramovitch, Jiang, Mendoza, Islam, Gonzalez, Mavroudis, Xu, Kumar, Goswami, Bugas, Heydari, Jeanplong, Jansen, Pinto, Apronti, Galal, Ze-An, Singh, Jiang, of~Arc~Xavier, Agarwal, Berkani, Zhang, Du, de~Oliveira~Junior, Malishev, Remy, Hartman, Tarver, Mensah, Loume, Morak, Habibi, Hoback, Cai, Gimenez, Montecillo, Łucki, Campbell, Sharma, Meer, Gul, Gonzalez, Alapont, Hoover, Chhablani, Vargus, Agarwal, Jiang, Patil, Outevsky, Scaria, Maheshwari, Dendane, Shukla, Cartwright, Bogdanov, Mündler, Möller, Arnaboldi, Thaman, Siddiqi, Saxena, Gupta, Fruhauff, Sherman, Vincze, Usawasutsakorn, Ler, Radhakrishnan, Enyekwe, Salauddin, Muzhen, Maksapetyan, Rossbach, Harjadi, Bahaloohoreh, Sparrow, Sidhu, Ali, Bian, Lai, Singer, Uro, Bateman, Sayed, Menshawy, Duclosel, Bezzi, Jain, Aaron, Tiryakioglu, Siddh, Krenek, Shah, Jin, Creighton,
  Peskoff, EL-Wasif, V, Richmond, McGowan, Patwardhan, Sun, Sun, Zubić, Sala, Ebert, Kaddour, Schottdorf, Wang, Petruzella, Meiburg, Medved, ElSheikh, Hebbar, Vaquero, Yang, Poulos, Zouhar, Bogdanik, Zhang, Sanz-Ros, Anugraha, Dai, Nhu, Wang, Demircali, Jia, Zhou, Wu, He, Chandok, Sinha, Luo, Le, Noyé, Perełkiewicz, Pantidis, Qi, Purohit, Parcalabescu, Nguyen, Winata, Ponti, Li, Dhole, Park, Abbondanza, Wang, Nayak, Caetano, Wong, del Rio-Chanona, Kondor, Francois, Chalstrey, Zsambok, Hoyer, Reddish, Hauser, Rodrigo-Ginés, Datta, Shepherd, Kamphuis, Zhang, Kim, Sun, Yao, Dernoncourt, Krishna, Rismanchian, Pu, Pinto, Wang, Shridhar, Overholt, Briia, Nguyen, David, Bartomeu, Pang, Wecker, Xiong, Li, Huber, Jaeger, Maddalena, Lù, Zhang, Beger, Kon, Li, Sanker, Yin, Liang, Zhang, Agrawal, Yifei, Zhang, Cai, Sonmez, Cozianu, Li, Slen, Yu, Park, Sarti, Briański, Stolfo, Nguyen, Zhang, Perlitz, Hernandez-Orallo, Li, Shabani, Juefei-Xu, Dhingra, Zohar, Nguyen, Pondaven, Yilmaz, Zhao, Jin, Jiang, Todoran, Han,
  Kreuer, Rabern, Plassart, Maggetti, Yap, Geirhos, Kean, Wang, Mollaei, Sun, Yin, Wang, Li, Chang, Wei, Bizeul, Wang, Arrais, Mukherjee, Chamorro-Padial, Liu, Qu, Guan, Bouyamourn, Wu, Plomecka, Chen, Tang, Deng, Subramanian, Xi, Chen, Zhang, Ren, Tu, Kim, Chen, Marjanović, Ha, Luczyna, Ma, Shen, Song, Zhang, Wang, Gendron, Xiao, Smucker, Weng, Lee, Ye, Ermon, Lopez-Miguel, Knights, Gitter, Park, Wei, Chen, Pai, Elkhanany, Lin, Siedler, Fang, Mishra, Zsolnai-Fehér, Jiang, Khan, Yuan, Jain, Lin, Peterson, Wang, Malusare, Tang, Gupta, Fosin, Kang, Dworakowska, Matsumoto, Zheng, Sewuster, Villanueva, Rannev, Chernyavsky, Chen, Banik, Racz, Dong, Wang, Bashmal, Gonçalves, Hu, Bar, Bohdal, Patlan, Dhuliawala, Geirhos, Wist, Kansal, Chen, Tire, Yücel, Christof, Singla, Song, Chen, Ge, Ponkshe, Park, Shi, Ma, Mak, Lai, Moulin, Cheng, Zhu, Zhang, Patil, Jha, Men, Wu, Zhang, Vieira, Aji, Chung, Mahfoud, Hoang, Sperzel, Hao, Meding, Xu, Kostakos, Manini, Liu, Toukmaji, Paek, Yu, Demircali, Sun, Dewerpe, Qin,
  Pflugfelder, Bailey, Morris, Heilala, Rosset, Yu, Chen, Yeo, Jain, Yang, Chigurupati, Chernyavsky, Reddy, Venugopalan, Batra, Park, Tran, Maximiano, Zhang, Liang, Shiyu, Xu, Pan, Suresh, Liu, Gulati, Zhang, Turchin, Bartlett, Scotese, Cao, Nattanmai, McKellips, Cheraku, Suhail, Luo, Deng, Luo, Zhang, Jindel, Paek, Halevy, Baranov, Liu, Avadhanam, Zhang, Cheng, Ma, Fu, Do, Lass, Yang, Sunkari, Bharath, Ai, Leung, Agrawal, Zhou, Chen, Kalpathi, Xu, Wang, Xiao, Maung, Lee, Yang, Yue, Zhao, Yoon, Sun, Singh, Luo, Peng, Osbey, Wang, Echeazu, Yang, Wu, Patel, Kulkarni, Sundarapandiyan, Zhang, Le, Nasim, Yalam, Kasamsetty, Samal, Yang, Sun, Shah, Saha, Zhang, Nguyen, Nagumalli, Wang, Zhou, Wu, Luo, Telluri, Yue, Wang, and Hendrycks]{hle}
Long Phan, Alice Gatti, Ziwen Han, Nathaniel Li, Josephina Hu, Hugh Zhang, Chen Bo~Calvin Zhang, Mohamed Shaaban, John Ling, Sean Shi, Michael Choi, Anish Agrawal, Arnav Chopra, Adam Khoja, Ryan Kim, Richard Ren, Jason Hausenloy, Oliver Zhang, Mantas Mazeika, Dmitry Dodonov, Tung Nguyen, Jaeho Lee, Daron Anderson, Mikhail Doroshenko, Alun~Cennyth Stokes, Mobeen Mahmood, Oleksandr Pokutnyi, Oleg Iskra, Jessica~P. Wang, John-Clark Levin, Mstyslav Kazakov, Fiona Feng, Steven~Y. Feng, Haoran Zhao, Michael Yu, Varun Gangal, Chelsea Zou, Zihan Wang, Serguei Popov, Robert Gerbicz, Geoff Galgon, Johannes Schmitt, Will Yeadon, Yongki Lee, Scott Sauers, Alvaro Sanchez, Fabian Giska, Marc Roth, Søren Riis, Saiteja Utpala, Noah Burns, Gashaw~M. Goshu, Mohinder~Maheshbhai Naiya, Chidozie Agu, Zachary Giboney, Antrell Cheatom, Francesco Fournier-Facio, Sarah-Jane Crowson, Lennart Finke, Zerui Cheng, Jennifer Zampese, Ryan~G. Hoerr, Mark Nandor, Hyunwoo Park, Tim Gehrunger, Jiaqi Cai, Ben McCarty, Alexis~C Garretson, Edwin
  Taylor, Damien Sileo, Qiuyu Ren, Usman Qazi, Lianghui Li, Jungbae Nam, John~B. Wydallis, Pavel Arkhipov, Jack Wei~Lun Shi, Aras Bacho, Chris~G. Willcocks, Hangrui Cao, Sumeet Motwani, Emily de~Oliveira~Santos, Johannes Veith, Edward Vendrow, Doru Cojoc, Kengo Zenitani, Joshua Robinson, Longke Tang, Yuqi Li, Joshua Vendrow, Natanael~Wildner Fraga, Vladyslav Kuchkin, Andrey~Pupasov Maksimov, Pierre Marion, Denis Efremov, Jayson Lynch, Kaiqu Liang, Aleksandar Mikov, Andrew Gritsevskiy, Julien Guillod, Gözdenur Demir, Dakotah Martinez, Ben Pageler, Kevin Zhou, Saeed Soori, Ori Press, Henry Tang, Paolo Rissone, Sean~R. Green, Lina Brüssel, Moon Twayana, Aymeric Dieuleveut, Joseph~Marvin Imperial, Ameya Prabhu, Jinzhou Yang, Nick Crispino, Arun Rao, Dimitri Zvonkine, Gabriel Loiseau, Mikhail Kalinin, Marco Lukas, Ciprian Manolescu, Nate Stambaugh, Subrata Mishra, Tad Hogg, Carlo Bosio, Brian~P Coppola, Julian Salazar, Jaehyeok Jin, Rafael Sayous, Stefan Ivanov, Philippe Schwaller, Shaipranesh Senthilkuma,
  Andres~M Bran, Andres Algaba, Kelsey~Van den Houte, Lynn Van~Der Sypt, Brecht Verbeken, David Noever, Alexei Kopylov, Benjamin Myklebust, Bikun Li, Lisa Schut, Evgenii Zheltonozhskii, Qiaochu Yuan, Derek Lim, Richard Stanley, Tong Yang, John Maar, Julian Wykowski, Martí Oller, Anmol Sahu, Cesare~Giulio Ardito, Yuzheng Hu, Ariel Ghislain~Kemogne Kamdoum, Alvin Jin, Tobias~Garcia Vilchis, Yuexuan Zu, Martin Lackner, James Koppel, Gongbo Sun, Daniil~S. Antonenko, Steffi Chern, Bingchen Zhao, Pierrot Arsene, Joseph~M Cavanagh, Daofeng Li, Jiawei Shen, Donato Crisostomi, Wenjin Zhang, Ali Dehghan, Sergey Ivanov, David Perrella, Nurdin Kaparov, Allen Zang, Ilia Sucholutsky, Arina Kharlamova, Daniil Orel, Vladislav Poritski, Shalev Ben-David, Zachary Berger, Parker Whitfill, Michael Foster, Daniel Munro, Linh Ho, Shankar Sivarajan, Dan~Bar Hava, Aleksey Kuchkin, David Holmes, Alexandra Rodriguez-Romero, Frank Sommerhage, Anji Zhang, Richard Moat, Keith Schneider, Zakayo Kazibwe, Don Clarke, Dae~Hyun Kim,
  Felipe~Meneguitti Dias, Sara Fish, Veit Elser, Tobias Kreiman, Victor Efren~Guadarrama Vilchis, Immo Klose, Ujjwala Anantheswaran, Adam Zweiger, Kaivalya Rawal, Jeffery Li, Jeremy Nguyen, Nicolas Daans, Haline Heidinger, Maksim Radionov, Václav Rozhoň, Vincent Ginis, Christian Stump, Niv Cohen, Rafał Poświata, Josef Tkadlec, Alan Goldfarb, Chenguang Wang, Piotr Padlewski, Stanislaw Barzowski, Kyle Montgomery, Ryan Stendall, Jamie Tucker-Foltz, Jack Stade, T.~Ryan Rogers, Tom Goertzen, Declan Grabb, Abhishek Shukla, Alan Givré, John~Arnold Ambay, Archan Sen, Muhammad~Fayez Aziz, Mark~H Inlow, Hao He, Ling Zhang, Younesse Kaddar, Ivar Ängquist, Yanxu Chen, Harrison~K Wang, Kalyan Ramakrishnan, Elliott Thornley, Antonio Terpin, Hailey Schoelkopf, Eric Zheng, Avishy Carmi, Ethan D.~L. Brown, Kelin Zhu, Max Bartolo, Richard Wheeler, Martin Stehberger, Peter Bradshaw, JP~Heimonen, Kaustubh Sridhar, Ido Akov, Jennifer Sandlin, Yury Makarychev, Joanna Tam, Hieu Hoang, David~M. Cunningham, Vladimir Goryachev,
  Demosthenes Patramanis, Michael Krause, Andrew Redenti, David Aldous, Jesyin Lai, Shannon Coleman, Jiangnan Xu, Sangwon Lee, Ilias Magoulas, Sandy Zhao, Ning Tang, Michael~K. Cohen, Orr Paradise, Jan~Hendrik Kirchner, Maksym Ovchynnikov, Jason~O. Matos, Adithya Shenoy, Michael Wang, Yuzhou Nie, Anna Sztyber-Betley, Paolo Faraboschi, Robin Riblet, Jonathan Crozier, Shiv Halasyamani, Shreyas Verma, Prashant Joshi, Eli Meril, Ziqiao Ma, Jérémy Andréoletti, Raghav Singhal, Jacob Platnick, Volodymyr Nevirkovets, Luke Basler, Alexander Ivanov, Seri Khoury, Nils Gustafsson, Marco Piccardo, Hamid Mostaghimi, Qijia Chen, Virendra Singh, Tran~Quoc Khánh, Paul Rosu, Hannah Szlyk, Zachary Brown, Himanshu Narayan, Aline Menezes, Jonathan Roberts, William Alley, Kunyang Sun, Arkil Patel, Max Lamparth, Anka Reuel, Linwei Xin, Hanmeng Xu, Jacob Loader, Freddie Martin, Zixuan Wang, Andrea Achilleos, Thomas Preu, Tomek Korbak, Ida Bosio, Fereshteh Kazemi, Ziye Chen, Biró Bálint, Eve J.~Y. Lo, Jiaqi Wang, Maria Inês~S.
  Nunes, Jeremiah Milbauer, M~Saiful Bari, Zihao Wang, Behzad Ansarinejad, Yewen Sun, Stephane Durand, Hossam Elgnainy, Guillaume Douville, Daniel Tordera, George Balabanian, Hew Wolff, Lynna Kvistad, Hsiaoyun Milliron, Ahmad Sakor, Murat Eron, Andrew Favre~D. O., Shailesh Shah, Xiaoxiang Zhou, Firuz Kamalov, Sherwin Abdoli, Tim Santens, Shaul Barkan, Allison Tee, Robin Zhang, Alessandro Tomasiello, G.~Bruno~De Luca, Shi-Zhuo Looi, Vinh-Kha Le, Noam Kolt, Jiayi Pan, Emma Rodman, Jacob Drori, Carl~J Fossum, Niklas Muennighoff, Milind Jagota, Ronak Pradeep, Honglu Fan, Jonathan Eicher, Michael Chen, Kushal Thaman, William Merrill, Moritz Firsching, Carter Harris, Stefan Ciobâcă, Jason Gross, Rohan Pandey, Ilya Gusev, Adam Jones, Shashank Agnihotri, Pavel Zhelnov, Mohammadreza Mofayezi, Alexander Piperski, David~K. Zhang, Kostiantyn Dobarskyi, Roman Leventov, Ignat Soroko, Joshua Duersch, Vage Taamazyan, Andrew Ho, Wenjie Ma, William Held, Ruicheng Xian, Armel~Randy Zebaze, Mohanad Mohamed, Julian~Noah Leser,
  Michelle~X Yuan, Laila Yacar, Johannes Lengler, Katarzyna Olszewska, Claudio~Di Fratta, Edson Oliveira, Joseph~W. Jackson, Andy Zou, Muthu Chidambaram, Timothy Manik, Hector Haffenden, Dashiell Stander, Ali Dasouqi, Alexander Shen, Bita Golshani, David Stap, Egor Kretov, Mikalai Uzhou, Alina~Borisovna Zhidkovskaya, Nick Winter, Miguel~Orbegozo Rodriguez, Robert Lauff, Dustin Wehr, Colin Tang, Zaki Hossain, Shaun Phillips, Fortuna Samuele, Fredrik Ekström, Angela Hammon, Oam Patel, Faraz Farhidi, George Medley, Forough Mohammadzadeh, Madellene Peñaflor, Haile Kassahun, Alena Friedrich, Rayner~Hernandez Perez, Daniel Pyda, Taom Sakal, Omkar Dhamane, Ali~Khajegili Mirabadi, Eric Hallman, Kenchi Okutsu, Mike Battaglia, Mohammad Maghsoudimehrabani, Alon Amit, Dave Hulbert, Roberto Pereira, Simon Weber, Handoko, Anton Peristyy, Stephen Malina, Mustafa Mehkary, Rami Aly, Frank Reidegeld, Anna-Katharina Dick, Cary Friday, Mukhwinder Singh, Hassan Shapourian, Wanyoung Kim, Mariana Costa, Hubeyb Gurdogan, Harsh
  Kumar, Chiara Ceconello, Chao Zhuang, Haon Park, Micah Carroll, Andrew~R. Tawfeek, Stefan Steinerberger, Daattavya Aggarwal, Michael Kirchhof, Linjie Dai, Evan Kim, Johan Ferret, Jainam Shah, Yuzhou Wang, Minghao Yan, Krzysztof Burdzy, Lixin Zhang, Antonio Franca, Diana~T. Pham, Kang~Yong Loh, Joshua Robinson, Abram Jackson, Paolo Giordano, Philipp Petersen, Adrian Cosma, Jesus Colino, Colin White, Jacob Votava, Vladimir Vinnikov, Ethan Delaney, Petr Spelda, Vit Stritecky, Syed~M. Shahid, Jean-Christophe Mourrat, Lavr Vetoshkin, Koen Sponselee, Renas Bacho, Zheng-Xin Yong, Florencia de~la Rosa, Nathan Cho, Xiuyu Li, Guillaume Malod, Orion Weller, Guglielmo Albani, Leon Lang, Julien Laurendeau, Dmitry Kazakov, Fatimah Adesanya, Julien Portier, Lawrence Hollom, Victor Souza, Yuchen~Anna Zhou, Julien Degorre, Yiğit Yalın, Gbenga~Daniel Obikoya, Rai, Filippo Bigi, M.~C. Boscá, Oleg Shumar, Kaniuar Bacho, Gabriel Recchia, Mara Popescu, Nikita Shulga, Ngefor~Mildred Tanwie, Thomas C.~H. Lux, Ben Rank, Colin
  Ni, Matthew Brooks, Alesia Yakimchyk, Huanxu, Liu, Stefano Cavalleri, Olle Häggström, Emil Verkama, Joshua Newbould, Hans Gundlach, Leonor Brito-Santana, Brian Amaro, Vivek Vajipey, Rynaa Grover, Ting Wang, Yosi Kratish, Wen-Ding Li, Sivakanth Gopi, Andrea Caciolai, Christian~Schroeder de~Witt, Pablo Hernández-Cámara, Emanuele Rodolà, Jules Robins, Dominic Williamson, Vincent Cheng, Brad Raynor, Hao Qi, Ben Segev, Jingxuan Fan, Sarah Martinson, Erik~Y. Wang, Kaylie Hausknecht, Michael~P. Brenner, Mao Mao, Christoph Demian, Peyman Kassani, Xinyu Zhang, David Avagian, Eshawn~Jessica Scipio, Alon Ragoler, Justin Tan, Blake Sims, Rebeka Plecnik, Aaron Kirtland, Omer~Faruk Bodur, D.~P. Shinde, Yan Carlos~Leyva Labrador, Zahra Adoul, Mohamed Zekry, Ali Karakoc, Tania C.~B. Santos, Samir Shamseldeen, Loukmane Karim, Anna Liakhovitskaia, Nate Resman, Nicholas Farina, Juan~Carlos Gonzalez, Gabe Maayan, Earth Anderson, Rodrigo De~Oliveira Pena, Elizabeth Kelley, Hodjat Mariji, Rasoul Pouriamanesh, Wentao Wu,
  Ross Finocchio, Ismail Alarab, Joshua Cole, Danyelle Ferreira, Bryan Johnson, Mohammad Safdari, Liangti Dai, Siriphan Arthornthurasuk, Isaac~C. McAlister, Alejandro~José Moyano, Alexey Pronin, Jing Fan, Angel Ramirez-Trinidad, Yana Malysheva, Daphiny Pottmaier, Omid Taheri, Stanley Stepanic, Samuel Perry, Luke Askew, Raúl Adrián~Huerta Rodríguez, Ali M.~R. Minissi, Ricardo Lorena, Krishnamurthy Iyer, Arshad~Anil Fasiludeen, Ronald Clark, Josh Ducey, Matheus Piza, Maja Somrak, Eric Vergo, Juehang Qin, Benjámin Borbás, Eric Chu, Jack Lindsey, Antoine Jallon, I.~M.~J. McInnis, Evan Chen, Avi Semler, Luk Gloor, Tej Shah, Marc Carauleanu, Pascal Lauer, Tran Đuc Huy, Hossein Shahrtash, Emilien Duc, Lukas Lewark, Assaf Brown, Samuel Albanie, Brian Weber, Warren~S. Vaz, Pierre Clavier, Yiyang Fan, Gabriel Poesia~Reis e~Silva, Long, Lian, Marcus Abramovitch, Xi~Jiang, Sandra Mendoza, Murat Islam, Juan Gonzalez, Vasilios Mavroudis, Justin Xu, Pawan Kumar, Laxman~Prasad Goswami, Daniel Bugas, Nasser Heydari,
  Ferenc Jeanplong, Thorben Jansen, Antonella Pinto, Archimedes Apronti, Abdallah Galal, Ng~Ze-An, Ankit Singh, Tong Jiang, Joan of~Arc~Xavier, Kanu~Priya Agarwal, Mohammed Berkani, Gang Zhang, Zhehang Du, Benedito~Alves de~Oliveira~Junior, Dmitry Malishev, Nicolas Remy, Taylor~D. Hartman, Tim Tarver, Stephen Mensah, Gautier~Abou Loume, Wiktor Morak, Farzad Habibi, Sarah Hoback, Will Cai, Javier Gimenez, Roselynn~Grace Montecillo, Jakub Łucki, Russell Campbell, Asankhaya Sharma, Khalida Meer, Shreen Gul, Daniel~Espinosa Gonzalez, Xavier Alapont, Alex Hoover, Gunjan Chhablani, Freddie Vargus, Arunim Agarwal, Yibo Jiang, Deepakkumar Patil, David Outevsky, Kevin~Joseph Scaria, Rajat Maheshwari, Abdelkader Dendane, Priti Shukla, Ashley Cartwright, Sergei Bogdanov, Niels Mündler, Sören Möller, Luca Arnaboldi, Kunvar Thaman, Muhammad~Rehan Siddiqi, Prajvi Saxena, Himanshu Gupta, Tony Fruhauff, Glen Sherman, Mátyás Vincze, Siranut Usawasutsakorn, Dylan Ler, Anil Radhakrishnan, Innocent Enyekwe, Sk~Md
  Salauddin, Jiang Muzhen, Aleksandr Maksapetyan, Vivien Rossbach, Chris Harjadi, Mohsen Bahaloohoreh, Claire Sparrow, Jasdeep Sidhu, Sam Ali, Song Bian, John Lai, Eric Singer, Justine~Leon Uro, Greg Bateman, Mohamed Sayed, Ahmed Menshawy, Darling Duclosel, Dario Bezzi, Yashaswini Jain, Ashley Aaron, Murat Tiryakioglu, Sheeshram Siddh, Keith Krenek, Imad~Ali Shah, Jun Jin, Scott Creighton, Denis Peskoff, Zienab EL-Wasif, Ragavendran~P V, Michael Richmond, Joseph McGowan, Tejal Patwardhan, Hao-Yu Sun, Ting Sun, Nikola Zubić, Samuele Sala, Stephen Ebert, Jean Kaddour, Manuel Schottdorf, Dianzhuo Wang, Gerol Petruzella, Alex Meiburg, Tilen Medved, Ali ElSheikh, S~Ashwin Hebbar, Lorenzo Vaquero, Xianjun Yang, Jason Poulos, Vilém Zouhar, Sergey Bogdanik, Mingfang Zhang, Jorge Sanz-Ros, David Anugraha, Yinwei Dai, Anh~N. Nhu, Xue Wang, Ali~Anil Demircali, Zhibai Jia, Yuyin Zhou, Juncheng Wu, Mike He, Nitin Chandok, Aarush Sinha, Gaoxiang Luo, Long Le, Mickaël Noyé, Michał Perełkiewicz, Ioannis Pantidis,
  Tianbo Qi, Soham~Sachin Purohit, Letitia Parcalabescu, Thai-Hoa Nguyen, Genta~Indra Winata, Edoardo~M. Ponti, Hanchen Li, Kaustubh Dhole, Jongee Park, Dario Abbondanza, Yuanli Wang, Anupam Nayak, Diogo~M. Caetano, Antonio A. W.~L. Wong, Maria del Rio-Chanona, Dániel Kondor, Pieter Francois, Ed~Chalstrey, Jakob Zsambok, Dan Hoyer, Jenny Reddish, Jakob Hauser, Francisco-Javier Rodrigo-Ginés, Suchandra Datta, Maxwell Shepherd, Thom Kamphuis, Qizheng Zhang, Hyunjun Kim, Ruiji Sun, Jianzhu Yao, Franck Dernoncourt, Satyapriya Krishna, Sina Rismanchian, Bonan Pu, Francesco Pinto, Yingheng Wang, Kumar Shridhar, Kalon~J. Overholt, Glib Briia, Hieu Nguyen, David, Soler Bartomeu, Tony~CY Pang, Adam Wecker, Yifan Xiong, Fanfei Li, Lukas~S. Huber, Joshua Jaeger, Romano~De Maddalena, Xing~Han Lù, Yuhui Zhang, Claas Beger, Patrick Tser~Jern Kon, Sean Li, Vivek Sanker, Ming Yin, Yihao Liang, Xinlu Zhang, Ankit Agrawal, Li~S. Yifei, Zechen Zhang, Mu~Cai, Yasin Sonmez, Costin Cozianu, Changhao Li, Alex Slen, Shoubin Yu,
  Hyun~Kyu Park, Gabriele Sarti, Marcin Briański, Alessandro Stolfo, Truong~An Nguyen, Mike Zhang, Yotam Perlitz, Jose Hernandez-Orallo, Runjia Li, Amin Shabani, Felix Juefei-Xu, Shikhar Dhingra, Orr Zohar, My~Chiffon Nguyen, Alexander Pondaven, Abdurrahim Yilmaz, Xuandong Zhao, Chuanyang Jin, Muyan Jiang, Stefan Todoran, Xinyao Han, Jules Kreuer, Brian Rabern, Anna Plassart, Martino Maggetti, Luther Yap, Robert Geirhos, Jonathon Kean, Dingsu Wang, Sina Mollaei, Chenkai Sun, Yifan Yin, Shiqi Wang, Rui Li, Yaowen Chang, Anjiang Wei, Alice Bizeul, Xiaohan Wang, Alexandre~Oliveira Arrais, Kushin Mukherjee, Jorge Chamorro-Padial, Jiachen Liu, Xingyu Qu, Junyi Guan, Adam Bouyamourn, Shuyu Wu, Martyna Plomecka, Junda Chen, Mengze Tang, Jiaqi Deng, Shreyas Subramanian, Haocheng Xi, Haoxuan Chen, Weizhi Zhang, Yinuo Ren, Haoqin Tu, Sejong Kim, Yushun Chen, Sara~Vera Marjanović, Junwoo Ha, Grzegorz Luczyna, Jeff~J. Ma, Zewen Shen, Dawn Song, Cedegao~E. Zhang, Zhun Wang, Gaël Gendron, Yunze Xiao, Leo Smucker, Erica
  Weng, Kwok~Hao Lee, Zhe Ye, Stefano Ermon, Ignacio~D. Lopez-Miguel, Theo Knights, Anthony Gitter, Namkyu Park, Boyi Wei, Hongzheng Chen, Kunal Pai, Ahmed Elkhanany, Han Lin, Philipp~D. Siedler, Jichao Fang, Ritwik Mishra, Károly Zsolnai-Fehér, Xilin Jiang, Shadab Khan, Jun Yuan, Rishab~Kumar Jain, Xi~Lin, Mike Peterson, Zhe Wang, Aditya Malusare, Maosen Tang, Isha Gupta, Ivan Fosin, Timothy Kang, Barbara Dworakowska, Kazuki Matsumoto, Guangyao Zheng, Gerben Sewuster, Jorge~Pretel Villanueva, Ivan Rannev, Igor Chernyavsky, Jiale Chen, Deepayan Banik, Ben Racz, Wenchao Dong, Jianxin Wang, Laila Bashmal, Duarte~V. Gonçalves, Wei Hu, Kaushik Bar, Ondrej Bohdal, Atharv~Singh Patlan, Shehzaad Dhuliawala, Caroline Geirhos, Julien Wist, Yuval Kansal, Bingsen Chen, Kutay Tire, Atak~Talay Yücel, Brandon Christof, Veerupaksh Singla, Zijian Song, Sanxing Chen, Jiaxin Ge, Kaustubh Ponkshe, Isaac Park, Tianneng Shi, Martin~Q. Ma, Joshua Mak, Sherwin Lai, Antoine Moulin, Zhuo Cheng, Zhanda Zhu, Ziyi Zhang, Vaidehi
  Patil, Ketan Jha, Qiutong Men, Jiaxuan Wu, Tianchi Zhang, Bruno~Hebling Vieira, Alham~Fikri Aji, Jae-Won Chung, Mohammed Mahfoud, Ha~Thi Hoang, Marc Sperzel, Wei Hao, Kristof Meding, Sihan Xu, Vassilis Kostakos, Davide Manini, Yueying Liu, Christopher Toukmaji, Jay Paek, Eunmi Yu, Arif~Engin Demircali, Zhiyi Sun, Ivan Dewerpe, Hongsen Qin, Roman Pflugfelder, James Bailey, Johnathan Morris, Ville Heilala, Sybille Rosset, Zishun Yu, Peter~E. Chen, Woongyeong Yeo, Eeshaan Jain, Ryan Yang, Sreekar Chigurupati, Julia Chernyavsky, Sai~Prajwal Reddy, Subhashini Venugopalan, Hunar Batra, Core~Francisco Park, Hieu Tran, Guilherme Maximiano, Genghan Zhang, Yizhuo Liang, Hu~Shiyu, Rongwu Xu, Rui Pan, Siddharth Suresh, Ziqi Liu, Samaksh Gulati, Songyang Zhang, Peter Turchin, Christopher~W. Bartlett, Christopher~R. Scotese, Phuong~M. Cao, Aakaash Nattanmai, Gordon McKellips, Anish Cheraku, Asim Suhail, Ethan Luo, Marvin Deng, Jason Luo, Ashley Zhang, Kavin Jindel, Jay Paek, Kasper Halevy, Allen Baranov, Michael Liu,
  Advaith Avadhanam, David Zhang, Vincent Cheng, Brad Ma, Evan Fu, Liam Do, Joshua Lass, Hubert Yang, Surya Sunkari, Vishruth Bharath, Violet Ai, James Leung, Rishit Agrawal, Alan Zhou, Kevin Chen, Tejas Kalpathi, Ziqi Xu, Gavin Wang, Tyler Xiao, Erik Maung, Sam Lee, Ryan Yang, Roy Yue, Ben Zhao, Julia Yoon, Sunny Sun, Aryan Singh, Ethan Luo, Clark Peng, Tyler Osbey, Taozhi Wang, Daryl Echeazu, Hubert Yang, Timothy Wu, Spandan Patel, Vidhi Kulkarni, Vijaykaarti Sundarapandiyan, Ashley Zhang, Andrew Le, Zafir Nasim, Srikar Yalam, Ritesh Kasamsetty, Soham Samal, Hubert Yang, David Sun, Nihar Shah, Abhijeet Saha, Alex Zhang, Leon Nguyen, Laasya Nagumalli, Kaixin Wang, Alan Zhou, Aidan Wu, Jason Luo, Anwith Telluri, Summer Yue, Alexandr Wang, and Dan Hendrycks.
\newblock Humanity's last exam, 2025.
\newblock URL \url{https://arxiv.org/abs/2501.14249}.

\bibitem[Rein et~al.(2024)Rein, Hou, Stickland, Petty, Pang, Dirani, Michael, and Bowman]{gpqa}
David Rein, Betty~Li Hou, Asa~Cooper Stickland, Jackson Petty, Richard~Yuanzhe Pang, Julien Dirani, Julian Michael, and Samuel~R. Bowman.
\newblock {GPQA}: A graduate-level google-proof q\&a benchmark.
\newblock In \emph{First Conference on Language Modeling}, 2024.
\newblock URL \url{https://openreview.net/forum?id=Ti67584b98}.

\bibitem[Sharma(2025)]{openevolve}
Asankhaya Sharma.
\newblock Openevolve: an open-source evolutionary coding agent, 2025.
\newblock URL \url{https://github.com/codelion/openevolve}.

\bibitem[Shojaee et~al.(2025)Shojaee, Nguyen, Meidani, Farimani, Doan, and Reddy]{llm-srbench}
Parshin Shojaee, Ngoc-Hieu Nguyen, Kazem Meidani, Amir~Barati Farimani, Khoa~D Doan, and Chandan~K Reddy.
\newblock Llm-srbench: A new benchmark for scientific equation discovery with large language models.
\newblock \emph{arXiv preprint arXiv:2504.10415}, 2025.

\bibitem[Starace et~al.(2025)Starace, Jaffe, Sherburn, Aung, Chan, Maksin, Dias, Mays, Kinsella, Thompson, Heidecke, Glaese, and Patwardhan]{paperbench}
Giulio Starace, Oliver Jaffe, Dane Sherburn, James Aung, Jun~Shern Chan, Leon Maksin, Rachel Dias, Evan Mays, Benjamin Kinsella, Wyatt Thompson, Johannes Heidecke, Amelia Glaese, and Tejal Patwardhan.
\newblock Paperbench: Evaluating ai's ability to replicate ai research, 2025.
\newblock URL \url{https://arxiv.org/abs/2504.01848}.

\bibitem[Wang et~al.(2024)Wang, Ma, Zhang, Ni, Chandra, Guo, Ren, Arulraj, He, Jiang, Li, Ku, Wang, Zhuang, Fan, Yue, and Chen]{mmlu-pro}
Yubo Wang, Xueguang Ma, Ge~Zhang, Yuansheng Ni, Abhranil Chandra, Shiguang Guo, Weiming Ren, Aaran Arulraj, Xuan He, Ziyan Jiang, Tianle Li, Max Ku, Kai Wang, Alex Zhuang, Rongqi Fan, Xiang Yue, and Wenhu Chen.
\newblock Mmlu-pro: A more robust and challenging multi-task language understanding benchmark, 2024.
\newblock URL \url{https://arxiv.org/abs/2406.01574}.

\bibitem[Zhao et~al.(2025)Zhao, Magka, Jiang, Li, Raileanu, Shavrina, Gagnon-Audet, Niu, Sodhani, Shvartsman, Lupu, Lupidi, Toledo, Hambardzumyan, Josifoski, Foster, Cipolina-Kun, Charnalia, Dunfield, Miller, Aodha, Foerster, and Bachrach]{automated-speedrunning}
Bingchen Zhao, Despoina Magka, Minqi Jiang, Xian Li, Roberta Raileanu, Tatiana Shavrina, Jean-Christophe Gagnon-Audet, Kelvin Niu, Shagun Sodhani, Michael Shvartsman, Andrei Lupu, Alisia Lupidi, Edan Toledo, Karen Hambardzumyan, Martin Josifoski, Thomas Foster, Lucia Cipolina-Kun, Abhishek Charnalia, Derek Dunfield, Alexander~H. Miller, Oisin~Mac Aodha, Jakob Foerster, and Yoram Bachrach.
\newblock The automated llm speedrunning benchmark: Reproducing nanogpt improvements, 2025.
\newblock URL \url{https://arxiv.org/abs/2506.22419}.

\bibitem[Zhong et~al.(2023)Zhong, Cui, Guo, Liang, Lu, Wang, Saied, Chen, and Duan]{agi-eval}
Wanjun Zhong, Ruixiang Cui, Yiduo Guo, Yaobo Liang, Shuai Lu, Yanlin Wang, Amin Saied, Weizhu Chen, and Nan Duan.
\newblock Agieval: A human-centric benchmark for evaluating foundation models, 2023.
\newblock URL \url{https://arxiv.org/abs/2304.06364}.

\end{thebibliography}
\appendix
\end{document}